\documentclass[lettersize,journal]{IEEEtran}
\usepackage{amsmath,amsfonts}
\usepackage{algorithmic}
\usepackage{algorithm}
\usepackage{array}
\usepackage[caption=false,font=normalsize,labelfont=sf,textfont=sf]{subfig}
\usepackage{textcomp}
\usepackage{stfloats}
\usepackage{url}
\usepackage{verbatim}
\usepackage{graphicx}
 \usepackage{multirow}

\usepackage[colorlinks=true, citecolor=blue, linkcolor=black, urlcolor=blue]{hyperref}
\usepackage{cite}
\def\BibTeX{{\rm B\kern-.05em{\sc i\kern-.025em b}\kern-.08em
    T\kern-.1667em\lower.7ex\hbox{E}\kern-.125emX}}
\begin{document}
\title{Tube Loss based Deep Networks For Improving the Probabilistic Forecasting of Wind Speed}
\author{Pritam Anand, \IEEEmembership{Fellow, IEEE}, Aadesh Minz, and Asish Joel.
\thanks{ This work was funded by the Smart Energy Learning Center at DA-IICT, Gandhinagar through Grant no.- CSR-25/BSES/A6-PA/SELC. }
\thanks{ Pritam Anand, Aadesh Minz and Asish Joel are with DA-IICT, Gandhinagar. (e-mail: pritam$\_$anand@daiict.ac.in) }
}

\markboth{}%
{Shell \MakeLowercase{\textit{et al.}}: A Sample Article Using IEEEtran.cls for IEEE Journals}


\maketitle

\begin{abstract}
Uncertainty Quantification (UQ) in wind speed forecasting is a critical challenge in wind power production due to the inherently volatile nature of wind. By quantifying the associated risks and returns, UQ supports more effective decision-making for grid operations and participation in the electricity market. In this paper, we design a sequence of deep learning based probabilistic forecasting methods by using the Tube loss function for wind speed forecasting. The Tube loss function is a  simple and model agnostic Prediction Interval (PI) estimation approach and can obtain the narrow PI with asymptotical coverage guarantees without any distribution assumption. Our deep probabilistic forecasting models effectively incorporate popular architectures such as LSTM, GRU, and TCN within the Tube loss framework. We further design a simple yet effective heuristic for tuning the $\delta$ parameter of the Tube loss function so that our deep forecasting models obtain the narrower PI without compromising its calibration ability.  We have considered three wind datasets, containing the hourly recording of the wind speed, collected from three distinct location namely Jaisalmer, Los Angeles and San  Fransico.  Our numerical results demonstrate that the proposed deep forecasting models produce more reliable and narrower PIs compared to recently developed probabilistic wind forecasting methods. 
\end{abstract}


\section{Introduction}
\label{sec:introduction}
\subsection{Background}
\IEEEPARstart{I}{t} is well known that fossil fuels cannot meet our  future energy needs because they are rapidly depleting and pose significant environmental risks. Wind energy is one of the fastest-growing clean sources of renewable energy, gaining sufficient global attention due to its unlimited availability and minimal environmental impact. 
In 2023, new global wind power capacity is going to surpass 100 GW for the first time \cite{council2017gwec}. The Global Wind Energy Council (GWEC) estimates the wind power capacity will reach to 143 GW by the end of this decade\cite{council2017gwec}. However, the high volatility and uncertainty present significant  challenges for maintaining the safe and stable operation of power systems, scheduling turbine maintenance, and integrating with the power grid \cite{heng2022probabilistic}\cite{afrasiabi2020advanced}. Therefore, the effective wind speed modelling and prediction plays a crucial role in overall wind power generation process.

There are plethora of methods and modeling techniques in the literature for wind speed forecasting using the historical wind data. But, due to highly random and chaotic nature of wind speed, their forecasting always involves a significant degree of uncertainty, which may lead to poor decisions in energy system management \cite{heng2022probabilistic}. Therefore, quantifying these uncertainties is crucial to mitigate the risks associated with various decisions related with wind energy management \cite{lin2018multi} such as  power system reserve setting \cite{lin2017multi} , unit commitment \cite{zhao2014expected}  and market trading \cite{wan2016pareto}.

\subsection{Overview of the previous work}
Given the time-series data, the probabilistic forecasting models obtain a pair of functions such that the future observation  would lie in the tube constructed by the estimated pair of functions with a given confidence $1-\alpha$.  Ultimately, the estimated pair of functions represent two different conditional quantiles \cite{koenker2005quantile} of the predictive distribution, such that their difference corresponds to the given confidence. Probabilistic forecasting is essentially a problem of estimating Prediction Interval (PI) with time-series data with a given target calibration $1-\alpha$.

There are several methods for probabilistic forecasting in literature but, they can be easily divided into two category on the basis of their initial assumption. There are a number of models which apriory assumes the predictive distribution of the data and estimate the parameters of the assumed distribution efficiently for obtaining the probabilistic forecast. These models are  known as \textit{Parametric or Distribution based} models. 

In  \cite{khosravi2012wind}  and \cite{khosravi2014optimized}, Khosravi et al. have used the Mean-Variance  Estimation (MVE) method for probabilistic forecasting of wind  power. They assume that the marginal distribution of target values follows Gaussian distribution and use a Neural Network (NN) to estimate the mean and variance of this distribution. 
Afrasiabi et al. have used the mixture of Gaussian density  for modeling the predictive distribution by stacking different deep neural architectures \cite{afrasiabi2020advanced}. In \cite{kou2013sparse} and \cite{scheuerer2015probabilistic}, authors realizes the  need of modeling the predictive distribution with non-Gaussian distribution for wind speed forecasting.  
However, the prior assumption of predictive distribution in these models are not sufficient to obtain the consistent performance across different variety of wind data.

In distribution-free setting, the wind probabilistic forecasting methods can be mainly divided into three major groups. 
\begin{enumerate}
\item [(1)] One group of wind models aims to estimate the overall predictive distribution for probabilistic forecasting. Many of these models, such as those by Bessa et al. (2012) and Khorramdel et al. (2018), use variants of kernel density estimation to estimate the wind density function. Probabilistic forecasts are then obtained by calculating the quantiles of the wind distribution function.

\item [(2)] The second group of wind models directly estimates the pair of conditional quantile functions for obtaining probabilistic forecast by minimizing 
the pinball loss function \cite{koenker2005quantile}. These models have gained popularity among researchers for modeling wind power due to its simplicity and consistent performance. Some of the noteworthy variants are \cite{heng2022probabilistic} \cite{cui2017wind}  \cite{hu2020novel} \cite{wan2016direct} \cite{yu2021regional} \cite{zhu2022wind}  \cite{cui2022ensemble} \cite{zhu2024large}.
One of the notable limitation of this approach in the deep learning application is that it requires the training of pair of deep learning models one by one for targeting the estimation of  pair of quantiles. 

\item[(3)] The third group of wind models minimize a loss function to simultaneously and directly estimate the pair of functions for probabilistic forecasting. As compared to quantile based deep forecasting models, they have two clear advantages. The first advantage is that these models require to be trained once and can directly estimate the both bound functions simultaneously. The second advantage is that they also allow the minimization of the width of  the PI tube in their optimization problem as they are estimating the bound functions of the PI simultaneously. The Lower Upper Bound Estimation (LUBE) \cite{khosravi2010lower} is one of the first popular model which designs loss function for PI estimation task. Later, the LUBE method was used with wind time-series data for probabilistic forecasting tasks in several literature . Some important of them are  \cite{wu2020probabilistic} \cite{li2021evolutionary} \cite{wang2020probabilistic} \cite{almutairi2022intelligent} \cite{chen2024novel} \cite{liu2021new}. But, one of the major problem  with the LUBE loss function is that it is a step like function and not differentiable, which  poses the challenge in  solving its optimization problem. This also limits its effective extension in deep learning framework, which generally require the gradient descent method based back-propagation for efficient network training. Taking motivation from this, Pearce et al., have proposed the Quality Driven (QD) loss function \cite{pearce2018high} for high quality PI estimation, which is similar to the loss function used in LUBE method. Though, QD loss function still remains step like function  but, it was approximated well by using the sigmoidal function.  It enables the QD loss based Neural Network to be efficiently trained using standard gradient descent methods, leading to better performance compared to LUBE-based neural networks. A sufficient empirical validations have been carried out for establishing the superiority of  the QD loss based Network over LUBE based Network in \cite{pearce2018high}. Also, Hu et al., have shown that the QD loss based deep networks outperform the LUBE Network for probabilistic forecasting of wind in \cite{hu2020new}.
\end{enumerate}

Now, we shall overview the architectures of deep networks used for the probabilistic forecasting of wind speed in brief. The overall architecture of deep networks used for wind probabilistic forecasting is the same as that used for wind point forecasting, except for the output layer. The neurons in the output layer of the network may vary depending on the approach used for probabilistic forecasting. Though, researchers have used the Feed Forward Neural Network, Deep Belief Network and other shallow architectures for wind speed forecasting but, recent literature emphasizes more upon the sequential deep networks for the wind forecasting tasks.  The Recurrent Neural Network (RNN), Gated Recurrent Unit (GRU) \cite{cho2014properties}, Long-Short Term Memory Network (LSTM) \cite{hochreiter1997long}, Temporal Convolution Network (TCN) \cite{lea2017temporal} are most popular deep learning architectures which are used in wind forecasting tasks. 
\subsection{Contributions}
In this paper, we design simple but, very promising  probabilistic forecasting algorithms using the novel Tube loss function \cite{anand2024tube} for wind speed without any distribution assumption. Before, listing our contribution, we provide the necessary details regarding the Tube loss function as follows.

The Tube loss function is originally developed for high quality PI estimation in distribution-free setting. For a given target confidence $1-\alpha$, the PI estimaton requires the estimation of the $q^{th}$ and $(1+q-\alpha)^{th}$ quantile functions for some $q \in (0, \alpha)$. Quantile-based PI models construct PI by independently estimating the lower and upper quantile bounds. This is achieved by solving two separate optimization problems, where each problem minimizes the pinball loss function, a loss specifically designed for quantile estimation. The first optimization targets the $q^{\text{th}}$ quantile function to estimate the lower bound, while the second focuses on the $(1 + q - \alpha)^{\text{th}}$ quantile function to estimate the upper bound. The Tube loss PI models simplify the overall PI estimation process by simultaneously  obtaining the both quantile bound of the PI by minimizing the simple Tube loss function. But, the main advantage of the Tube loss based PI models over the Quantile based PI models is their ability to explicitly minimize the width of the PI in their optimization problem. This often leads to narrower PI in Tube loss models, without compromising their ability to achieve the target calibration of $1 - \alpha$.  

In addition to this, the Tube loss function has also other appealing properties, which makes it one of the most natural choice for PI estimation tasks.  Unlike the LUBE and QD loss functions, the Tube loss function is differentiable (re Lebesgue measure) everywhere which makes it minimization handful with gradient descent method without using any approximation. Also, the minimizer of the Tube loss function asymptotically guarantees the target calibration $1-\alpha$. Further, the Tube loss function also facilitate the movement of the PI tube through its $r$ parameters. This enables Tube loss-based PI models to learn PI that pass through the denser regions of the data cloud, resulting in narrower PI, particularity, beneficial in case of asymmetric noise in data.

Now, we list the contribution of this paper as follows.
\begin{enumerate}
    \item We extend the idea of the Tube loss into the probabilistic forecasting within auto-regressive framework. The Tube loss probabilistic forecasting  method is model-agnostic  and can be used with any deep auto-regressive architecture.

\item We develop three Tube loss-based deep forecasting architectures using LSTM, GRU, and TCN models. Additionally, we design a simple yet effective heuristic for tuning the parameter of the Tube loss function within these architectures for obtaining the narrow PI while wind forecasting.

\item We have considered three wind datasets collected from three distinct
location namely Jaisalmer, Los Angeles and San Fransico. We comapre our Tube loss based probabilistic forecasting models with recently developed top performing complex deep forecasting architectures for wind speed including Quantile and QD loss based deep forecasting models, Deep AR\cite{salinas2020deepar}, Mixture Density Network\cite{bishop1994mixture} and pretrained Time-GPT model \cite{garza2023timegpt} on considered datasets. Numerical results conclude that the Tube loss based deep forecasting architectures significantly outperform the  recent popular wind deep probabilistic forecasting models.       
\end{enumerate}

\subsection{Organization}
We have organized the rest of the paper as follows. Section \ref{2} details about the Tube loss function along with its appealing proprieties. In Section \ref{3}, we extend the Tube loss for time-series data for handling the probabilistic forecasting tasks and detailed the proposed methodology . Section \ref{4}  discusses the numerical results which show the efficacy of the Tube loss based deep learning models over existing and recent complex deep learning models developed for wind probabilistic forecasting task.

\section{Tube loss function}
\label{2}
Given training set $T=\{(x_i,y_i): x_i \in \mathbf{R}^n, y_i \in \mathbf{R}, i=1,2,...m\}$, sampled independently from pair of random variables $(X,Y)$, with target calibration $1-\alpha$, the PI estimation task is to estimate a pair of functions $(\mu_1,\mu_2)$ such that $P(\mu_1(X) \leq Y \leq \mu_2(X)|X) \geq 1-\alpha$.  In practice, the PI model aims to search the narrowest possible PI tube, subject to the constraint that the proportion of $y$ values lying within the PI tube is at least equal to the target coverage level $1-\alpha$. 

For given $u_1 > u_2$, we define the Tube loss function using
 \begin{eqnarray}
		  \rho_{1-\alpha}^{r}(u_2, u_1) =
		 \begin{cases}
                 (1-\alpha) u_2,  ~~\mbox{if}~~~ u_2  > 0,\\
			-\alpha u_2,  \mbox{~~~~~~~if}~u_2 \leq 0 ,u_1 \geq 0   \mbox{~and~~}  \nonumber \\~~~~~~~~~~~~~~~~~ {ru_2+(1-r)u_1} \geq 0,\\
			\alpha u_1, \mbox{~~~~~~~~~if}~u_2 \leq 0 ,u_1 \geq 0  \mbox{~and~~} \nonumber \\   ~~~~~~~~~~~~~~~~~ {ru_2+(1-r)u_1} < 0,\\
			-(1-\alpha) u_1, ~~ \mbox{if}~~ u_1  < 0, 
		\end{cases}\\
		\label{ppl1}
	\end{eqnarray}
 where $0 <r < 1$ is a user-defined parameter and ($u_2$, $u_1$) are errors, representing the deviations of $y$ values from the bounds of PI.  
 \begin{figure}[htp]
		\centering 
 {\includegraphics[width=0.7\linewidth, height = 0.35\linewidth ]{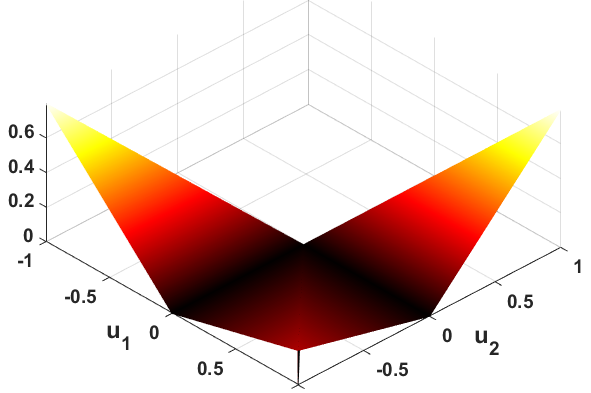}}   
  \caption{Tube loss function for $1-\alpha =0.9$}
  \label{tubeloss}
  \end{figure}
 
 For obtaining the estimates of the PI $[{\mu}_{1}(x),{\mu}_{2}(x)]$ with target calibration $1-\alpha$ and training set $T$,  we consider the $u_1 := y-\mu_1(x)$ and $u_2 := y-\mu_2(x)$  in the Tube loss function (\ref{ppl1}), which reduces it to 
 \begin{eqnarray}
		&\hspace{-140mm}  \rho_{1-\alpha}^{r} (y,\mu_{1}(x),\mu_{2}(x) ) = \nonumber \\
		 \begin{cases}
   t (y-\mu_2(x)),  \mbox{~~~~~~~~if}~~ y > \mu_2(x) . \\
   (1-t) (\mu_2(x)-y), \mbox{~if}~~~ \mu_1(x) \leq y \leq \mu_2(x)  
 \nonumber \\  &  \hspace{-40mm} \mbox{~and~~} y \geq {r\mu_2(x)+(1-r)\mu_1(x)}. \\
			(1-t) (y-\mu_1(x)),  \mbox{~if}~~~ \mu_1(x) \leq y\leq \mu_2(x)  \nonumber \\  &  \hspace{-40mm} \mbox{~~and~~} y < {r\mu_2(x)+(1-r)\mu_1(x)}.\\
			 t (\mu_1(x)-y), ~~~~~~~ ~~~~\mbox{if}~~~  y <  \mu_1(x).
		\end{cases} \\
		\label{pl1}
	\end{eqnarray}
 
 Let us  denote the  $(\hat{\mu}_{1}(x),\hat{\mu}_{2}(x))$ is the minimizer of average loss of the Training set $T=\{(x_i,y_i): x_i \in \mathbf{R}^n, y_i \in \mathbf{R}, i=1,2,...m\}$,  by the Tube loss  (\ref{pl1})  i,e. \[  \min_{(\mu_{1},\mu_{2})} \sum_{i=1}^{m} \rho_{t}^{r} (y_i, \mu_{1}(x_i),  \mu_{2}(x_i) ),\], then the  PI $[\hat{\mu}_{1}(x),\hat{\mu}_{2}(x)]$ guaranties to cover $1-\alpha$ fraction of $y$ values as $m \rightarrow \infty$. The  coverage obtained by PI with minimizer of the  Tube loss function  $(\hat{\mu}_{1}(x),\hat{\mu}_{2}(x))$  reaches to the target $1-\alpha$ asymptotically. The proof  of this is is detailed in \cite{anand2024tube}.

\begin{figure}[hbt!]
\centering
{ \includegraphics[width=0.7\linewidth, height = 0.5\linewidth ]{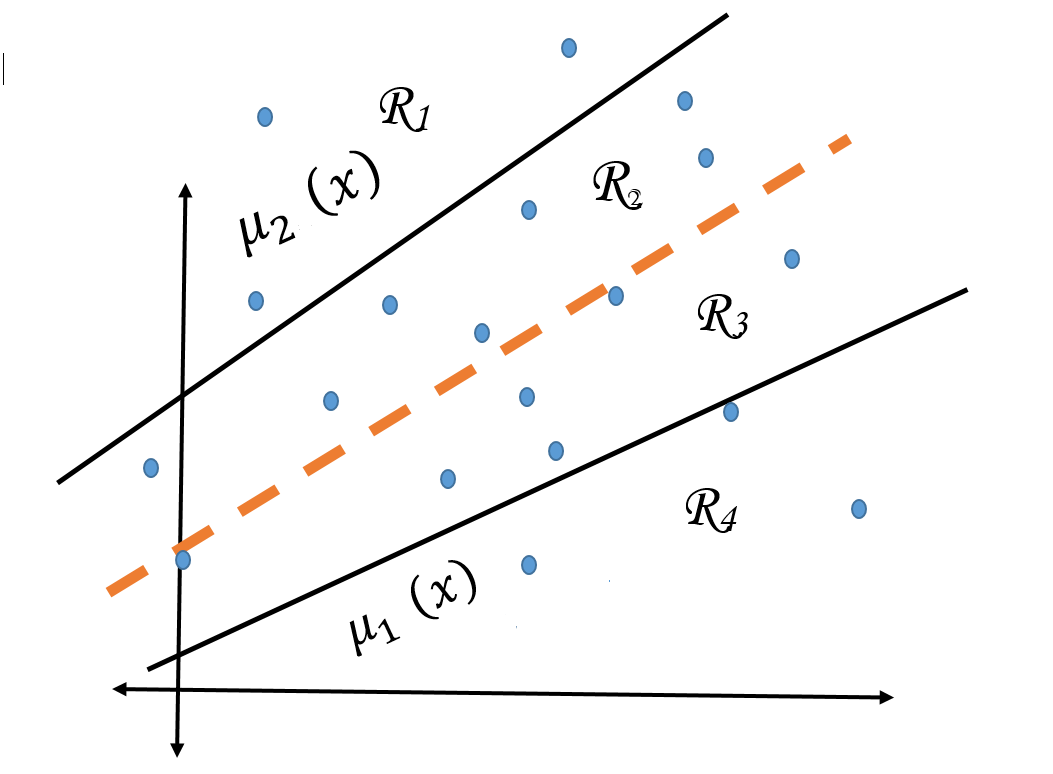}}  
\caption{Tube loss methodology:  red line represents the convex combination of ${\mu}_{1}(x)$ and ${\mu}_{2}(x)$, i,e., $r{\mu}_{1}(x)+(1-r){\mu}_{2}(x)$, $0<r<1$. Blue dots represent data points $(x_i,y_i)$, $i =1,2,,..,m$. For simplicity, both PI bounds  considered as the linear function of $x$.}
\label{hist2}
\end{figure}

 The Tube loss methodology divides the input-target space into four disjoint regions as shown in the Figure \ref{hist2}
such that $\frac{m_1}{m_2}$  and $\frac{m_3}{m_4} = \frac{1-t}{t}$  as $m \rightarrow \infty$, where $m_i$ is the number of data points lying in the  $\Re_i$ region for $i=1,2,3,4$ (See Figure \ref{hist2}).

The PI tube can be moved up or down  by varying the choice of parameter $r$. The default choice of $r$ parameter in the Tube loss is 0.5, which assumes the noise distribution is symmetric and  estimate the centered PI. But, when the distribution of $y\mid x$ is skewed, the centered PI may not cover the densest region of the data cloud and hence can obtain large PI width. In such cases, the PI tube can be moved up or down for obtaining the PI with lesser width.  A more detailed discussion and reasoning behind the $r$ parameter relating to the movement of PI tube in Tube loss method is provided in \cite{anand2024tube}.

 \section{ Probabilistic Forecasting of Wind Speed using Tube loss }
 \label{3}

  \begin{figure*}[h]
\centering
\includegraphics[width=15cm]{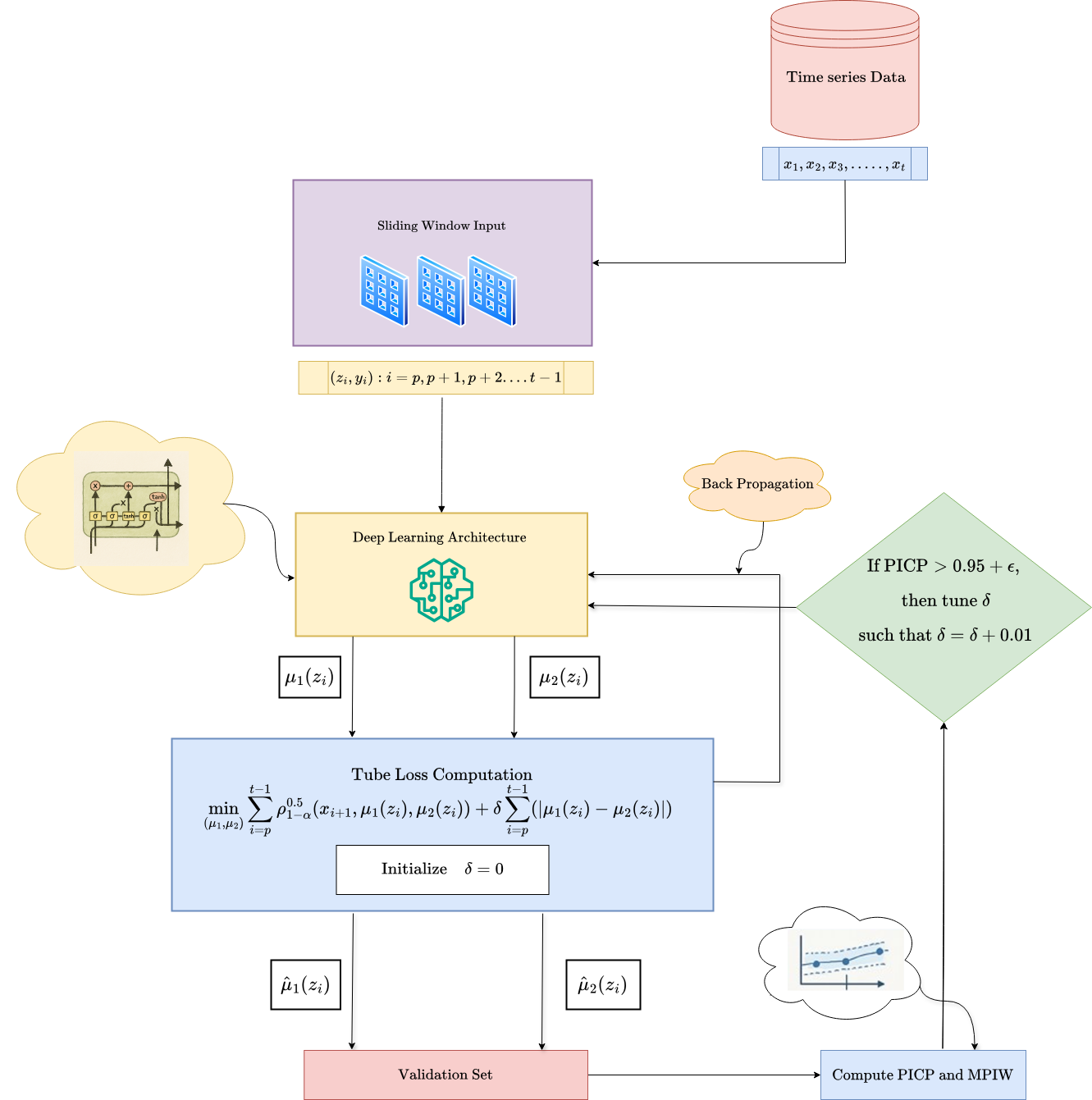}
\caption{Flow Diagram: Probabilistic forecasting with Tube loss based deep learning models}
\label{method_diag}
\end{figure*}

  Let us consider the time series wind speed observations $ (x_1,x_2,....,x_t)$, which are observed on $t$ different time stamps. If $p < t$ be the effective lag window, then auto-regressive models estimates the relationship between $z_i:= (x_{(i-p+1)},,....,x_i) $  and $x_{i+1}$ for $ i = p, p+1,...,t-1 $ using training set $T$ and use this relationship for forecasting the future observations. Apart from the Gaussian noise model, it can also consider the negative binomial noise model for probabilistic forecasting. 
  
 The task of probabilistic forecasting is to estimate the PI for $x_{i+1}$ given the input $z_{i}$ for $i \geq  t$. A deep probabilistic forecasting method  estimate the PI   $[\hat{\mu}_{1}(z_{i}),\hat{\mu}_{2}(z_{i})]$ , where $\hat{\mu}_{1}(z_{i})$ and $\hat{\mu}_{1}(z_{i})$ are estimated  quantile bound functions generated by  deep auto-regressive architecture such as LSTM, GRU or TCN.  For given target calibration $1-\alpha$, the $\hat{\mu}_{1}(z_{i})$ and $\hat{\mu}_{2}(z_{i})$ are the estimate of the $q^{th}$ and ${(1-\alpha-q)}^{th}$ quantiles of the  conditional distribution $(x_{i+1} \mid z_{i})$ for some $ q \in (0,\alpha)$. Distribution -free probabilistic forecasting methods directly estimate these quantile function without making any assumption regrading the conditional distribution $(x_{i+1} \mid z_{i})$.

In quantile based deep forecasting models, two separate deep auto-regressive  architecture is trained for estimating the  $q^{th}$ and ${(1-\alpha-q)}^{th}$ quantile functions independently for some $ q \in (0,\alpha)$. It makes the tunning and training of probabilistic models tedious and complex.   

The Tube loss based deep forecasting model simplifies the overall process of probabilistic forecasting and simultaneously estimate the both bound of PI by training a single model. The Tube loss based deep forecasting architecture contains the two nodes in its output layer which corresponds to the bounds of the PI, $\mu_1(z_{i})$ and $\mu_2(z_{i})$, and minimize the following problem
\begin{equation}
 \min_{(\mu_{1},\mu_{2})} \sum \limits_{i=p}^{t-1} \rho_{1-\alpha}^{r} (x_{i+1}, \mu_{1}(z_i),  \mu_{2}(z_i)) + \delta \sum_{i=p}^{t-1} (| \mu_{1}(z_i) -\mu_{2}(z_i)|)  \label{tube_prob}
\end{equation}
 The solution of this problem  i,e., $[({\mu}_{1},{\mu}_{2})]$, can be obtained by   back-propagating the loss (\ref{tube_prob})  through the entire  deep architecture using stochastic gradient descent method. After training, the PI estimate for the future observation $x_{t+1}$ can be obtained by $[\hat{\mu}_{1}(z_t),\hat{\mu}_{2}(z_t)]$.

 The Tube loss deep learning models obtain the both bounds of the PI simultaneously in probabilistic forecasting.  Unlike the Quantile based deep models, it facilitates to minimize the width of the PI tube  in the optimization problem against the empirical loss explicitly through user defined parameter $\delta$. It causes to reduce the width of the PI on test set significancy in practice. The $r$ parameter in the Tube loss problem (\ref{tube_prob}) enables the movement of the PI tube up and down in the probabilistic forecast while maintaining its calibration. 

 In  Figure (\ref{method_diag}), we have illustrated the overall flow (top to bottom) of methodology used by our Tube loss based deep learning models for probabilistic forecasting of wind speed.  At first, the time-series wind speed dataset is collected and last $30 \%$ observations were kept as test set. From training set, the last $10\%$ observations were used for the validation. Thereafter,  proper choice of the lag $p$ is obtained or tuned and sliding window technique is used to construct the features $z_i$ against the target $x_{i+1}$ in auto-regressive setting. 
 
We have set the target calibration $1-\alpha = 0.95$ for our all experiments. For evaluating the quality of the forecast, we have mainly used the two popular evaluation criteria , Prediction Interval Coverage Probability (PICP) and Mean Prediction Interval Width (MPIW).  The PICP counts the fraction of  $y$ values  lying inside of the estimated PI on test set. It should be equal or greater than the target calibration $1-\alpha$. In auto-regressive setting,  for test set $\{x_{t+1},x_{t+2},...,x_{t+N}\}$,  it can be computed by
\begin{equation}
\text{PICP:=} = \frac{1}{N} \sum_{i=1}^{N} \mathbb{I} \Big \{ x_{t+i} \in [\hat{\mu}_{1}(z_{t+i-1}),\hat{\mu}_{2}(z_{t+i-1})]. \Big \}
\end{equation}
 Here, $\mathbb{I}$ denotes the indicator function, which returns 1 if the condition is true, and 0 otherwise.
The MPIW for the given test set can be computed using the
\begin{equation}
  \mbox{MPIW:} =  \frac{1}{N} \sum \limits_{i=1}^{N} \{ \hat{\mu}_{2}(z_{t+i-1})-\hat{\mu}_{1}(z_{t+i-1}) \}
\end{equation}
   One another important evaluation metric used in recent wind literature for evaluating the probabilistic forecasting is the Continuous Rank Probability Score (CRPS) which quantify  goodness of the  estimates
   the conditional distribution  $(x_{i+1}\mid z_{i})$.   However, this evaluation metric is not relevant in  case of evaluating the quality of the probabilistic forecast obtained by the Tube loss  based deep learning  models as they directly obtains the two bounds of the PI explicitly without estimating the conditional distribution  $(x_{i+1}\mid z_{i})$. 
   
 As shown in the Figure (\ref{method_diag}), the  Tube loss deep forecasting architecture contains pair of nodes in the output layer which corresponds to the quantile bounds of the PI. Thereafter, the Tube loss objective (\ref{tube_prob}) is set to be minimized in which the empirical calibration (measured by the Tube loss) is tradded off against the width of the PI through the $\delta$ parameter. The parameter $\delta$ requires the efficient tuning for high quality PI. We design a simple but effective heuristic to adjust the value of the parameter $\delta$. At first, the deep learning architecture is trained by backpropagating the tube loss objective (\ref{tube_prob}) with $\delta =$0 and the PI obtained is evaluated in the validation set using PICP and MPIW. If the observed PICP on the validation set is significantly higher than the target calibration level of 0.95, there is an opportunity to reduce the MPIW value, since MPIW increases with PICP. In such cases, we increment the value of $\delta$ by 0.01 and retrain our deep forecasting architecture to achieve lower MPIW values. We shall refer this process as \textit{recalibration}. Through the repetitive \textit{recalibration}, the Tube loss based deep forecasting models not only achieve the target calibration but also, obtain the narrower PI. It causes them to provide actionable and precise forecast for effective  decision-making regarding the energy scheduling and grid management.
 
\section{Experimental Results}
\label{4}
We have selected three datasets collected from three different cities namely Jaisalmer,  Los Angeles and San Francisco. Jaisalmer, Los Angeles and San Francisco datasets contain the hourly recording of the wind speed  at 120 m for 8760 , 26304, 26304 hours respectively.  We have used the three popular deep auto-regressive models LSTM, GRU and TCN in the the Tube loss methodology to show its effectiveness. 

\subsection{Baseline Methods and Comparison Rule }
The Quantile Regression through pinball loss function is one of the most popular choice among researchers for probabilistic forecasting of wind speed. Different variants of the Neural and deep learning models utilizing the Quantile Regression approach for wind probabilistic forecasting are explored in  \cite{heng2022probabilistic} \cite{cui2017wind}  \cite{hu2020novel} \cite{wan2016direct} \cite{yu2021regional} \cite{zhu2022wind}  \cite{cui2022ensemble} \cite{zhu2024large}.  Also, the QD loss based deep architectures are used for probabilistic forecasting of wind in \cite{saeed2024short}  \cite{hu2020new}.   Therefore, we have 
 used the LSTM, GRU , TCN  architecture with the Quantile and QD loss approach as the baseline methods for comparisons.

 Recently, the Deep AR \cite{salinas2020deepar} model has also emerged as a popular method for probabilistic forecasting and is used in various time series AI applications, including wind power \cite{arora2022probabilistic}. Further,  multiple recent wind literature of probabilistic forecasting like \cite{yang2021improved} \cite{zhang2020improved} \cite{men2016short}   uses the variants of the Mixture Density Networks (MDN) model and show its efficacy over the existing methods. We have included the the Deep AR \cite{salinas2020deepar} and the MDN model for probabilistic forecasting of wind speed as baseline method in our experiments.
  In addition to this, we have also used the pre-trained Time-GPT model \cite{garza2023timegpt}, a foundation model for time series forecasting, as the baseline method for comparison.

  To compare the two probabilistic forecasting methods, we use a straightforward and intuitive rule. If both methods achieve a PCIP greater than the target value of 0.95, we consider the one with the smaller MPIW to be better. If neither method reaches the 0.95 target, then the method with a PCIP closer to 0.95 is preferred. In the remaining case, where only one method meets the target calibration of 0.95, that method is considered better than the one that does not.

\subsection{Numerical Results}
We present the performances of the Tube loss based deep learning models  and compare them with the recent baseline methods on Jaisalmer,  Los Angeles and San Francisco in Table \ref{tab:jaisalmer}, \ref{tab:losangeles} and \ref{tab:losangeles} respectively.

\begin{table}[h]
\centering
\caption{Performance Comparison of Different Models on Jaisalmer Dataset}
\label{tab:jaisalmer}
\begin{tabular}{|l|l|l|c|c|c|}
\hline
\textbf{Dataset} & \textbf{Model} & \textbf{Loss Function} & \textbf{PICP} & \textbf{MPIW} & \textbf{Time (s)} \\ \hline
\multirow{12}{*}{Jaisalmer} 
    & \multirow{3}{*}{LSTM} & \textbf{Tube} & \textbf{0.9589} & \textbf{3.697} & {18.86} \\
    &                       & QD            & 0.9689          & 4.47           & 28.33          \\
    &                       & Quantile      & 0.979           & 4.937          & 24.14          \\ \cline{2-6} 
    & \multirow{3}{*}{GRU}  & \textbf{Tube} & \textbf{0.955}  & \textbf{3.392} & {50.48} \\
    &                       & QD            & 0.9666          & 3.966          & 19.41          \\
    &                       & Quantile      & 0.9891          & 5.365          & 34.7           \\ \cline{2-6} 
    & \multirow{3}{*}{TCN}  & \textbf{Tube} & \textbf{0.9581} & \textbf{3.453} & {14.68} \\
    &                       & QD            & 0.9589          & 3.511          & 11.18          \\
    &                       & Quantile      & 0.9674          & 4.094          & 16.35          \\ \cline{2-6} 
    & \multicolumn{2}{l|}{MDN}              & 0.955           & 4.626          & 91.15          \\
    & \multicolumn{2}{l|}{Time GPT}         & 0.9417          & 10.608         & 3.62           \\
    & \multicolumn{2}{l|}{DEEPAR}           & 0.9969          & 6.3553         & 110.68         \\ \hline
\end{tabular}
\end{table}


\begin{table}[h]
\centering
\caption{Performance Comparison of Different Models on Los Angeles Dataset}
\label{tab:losangeles}
\begin{tabular}{|l|l|l|c|c|c|}
\hline
\textbf{Dataset} & \textbf{Model} & \textbf{Loss Function} & \textbf{PICP} & \textbf{MPIW} & \textbf{Time (s)} \\ \hline
\multirow{12}{*}{\shortstack{Los \\ Angeles}} 
    & \multirow{3}{*}{LSTM} & \textbf{Tube} & \textbf{0.9553} & \textbf{3.589} & {56.21}    \\
    &                       & QD            & 0.9467          & 3.471          & 70.9              \\
    &                       & Quantile      & 0.9547          & 4.001          & 80.81             \\ \cline{2-6} 
    & \multirow{3}{*}{GRU}  & {Tube} & {0.9586} & {3.737} & {143.19}   \\
    &                       & QD            & 0.9528          & 3.749          & 33.21             \\
    &                       & \textbf{Quantile}      &  \textbf{0.951}           & \textbf{3.618}          & 95.19             \\ \cline{2-6} 
    & \multirow{3}{*}{TCN}  & \textbf{Tube} & \textbf{0.9502} & \textbf{3.555} & {15.58}    \\
    &                       & \textbf{QD}            & \textbf{0.9507}          & \textbf{3.548}          & 44.25             \\
    &                       & Quantile      & 0.9416          & 3.779          & 34.54             \\ \cline{2-6} 
    & \multicolumn{2}{l|}  {MDN}              & \textbf{0.956}           & \textbf{3.487}          & 527.84            \\
    & \multicolumn{2}{l|}{Time GPT}         & 0.9517          & 6.519          & 3.28              \\
    & \multicolumn{2}{l|}{DEEPAR}           & 0.9773          & 4.5008         & 314.5             \\ \hline
\end{tabular}
\end{table}

\begin{table}[!ht]
\centering
\caption{Performance Comparison of Different Models on San Francisco Dataset}
\label{tab:sanfrancisco}
\begin{tabular}{|l|l|l|c|c|c|}
\hline
\textbf{Dataset} & \textbf{Model} & \textbf{Loss Function} & \textbf{PICP} & \textbf{MPIW} & \textbf{Time (s)} \\ \hline
\multirow{12}{*}{\shortstack{San\\Francisco}}
    & \multirow{3}{*}{LSTM} & \textbf{Tube} & \textbf{0.9561} & \textbf{4.627} & {126.6}  \\
    &                       & QD            & 0.9734          & 6.043          & 74.01           \\
    &                       & Quantile      & 0.9594          & 5.407          & 56.81           \\ \cline{2-6} 
    & \multirow{3}{*}{GRU}  & \textbf{Tube} & \textbf{0.9507} & \textbf{4.857} & {103.66} \\
    &                       & QD            & 0.9724          & 6.309          & 46.76           \\
    &                       & Quantile      & 0.951           & 4.955          & 101.26          \\ \cline{2-6} 
    & \multirow{3}{*}{TCN}  & \textbf{Tube} & \textbf{0.9543} & \textbf{4.56}  & {21.67}  \\
    &                       & QD            & 0.9505          & 5.49           & 22.63           \\
    &                       & Quantile      & 0.9428          & 4.908          & 27.69           \\ \cline{2-6} 
    & \multicolumn{2}{l|}{MDN}              & 0.949           & 4.62           & 511.7           \\
    & \multicolumn{2}{l|}{Time GPT}         & 0.9357          & 11.235         & 4.23            \\
    & \multicolumn{2}{l|}{DEEPAR}           & 0.9855          & 6.2023         & 313.45          \\ \hline
\end{tabular}
\end{table}

\subsection{Analysis}
We present a detailed analysis of the numerical results obtained by the different probabilistic forecasting models in in Table \ref{tab:jaisalmer}, \ref{tab:losangeles} and \ref{tab:losangeles}.

\begin{figure*}
\centering
\includegraphics[width=0.90\linewidth,height = 0.2\linewidth]{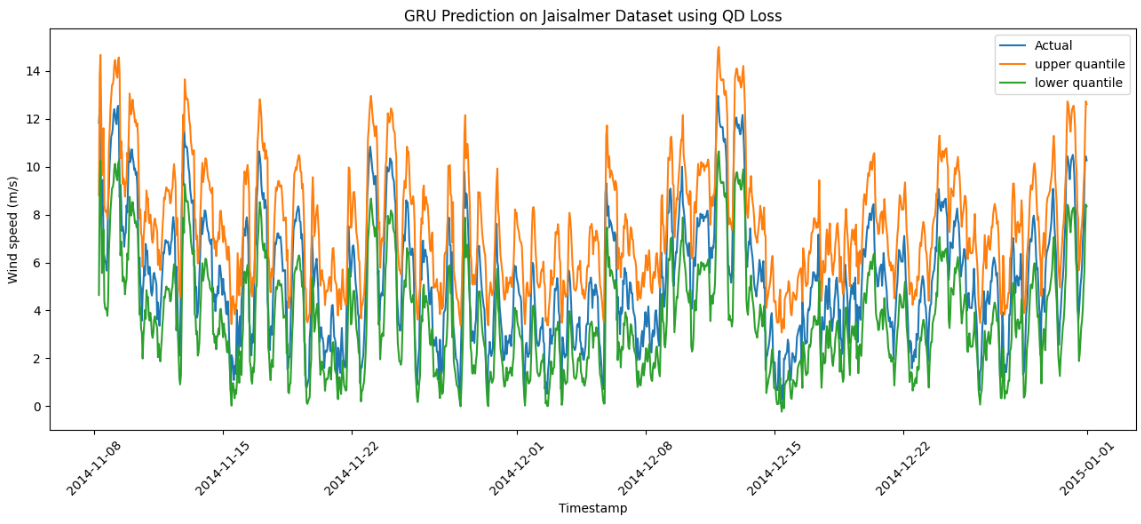}
\caption{Plot of GRU-Tube on Jaisalmer Dataset}
\label{fig:GRU_Tube}
\end{figure*}

\textbf{Jaisalmer:-} On the Jaisalmer dataset, Tube loss based methods consistently achieve the target calibration level of $0.95$ and outperforms competing methods. Notably, the GRU + Tube model achieves a PICP of $0.955$ with an MPIW of $3.392$, outperforming GRU + Quantile (PICP = $0.9891$, MPIW = $5.365$) by producing intervals that are approximately $37\%$ narrower. Similarly, the TCN + Tube model attains a PICP of $0.9581$ and an MPIW of $3.453$, which is about $16\%$ narrower than TCN + Quantile (MPIW = $4.094$). In the case of LSTM, the Tube loss approach improves the MPIW of the Quantile method by $25.12\%$. Furthermore, across all tested deep architectures, the Tube loss method consistently outperforms the QD loss model. The DeepAR and MDN obtain very wide PI as compared to the distribution-free probabilistic forecasting models. 
\begin{table}[h]
    \centering
    \caption{Jaisalmer Dataset Rankings}
    \label{tab:jaisalmer_ranking}
    \begin{tabular}{|c|l|c|c|c|}
        \hline
        \textbf{Rank} & \textbf{Model} & \textbf{PICP} & \textbf{MPIW} &\textbf{MPIW/PICP}\\
        \hline
        1 & GRU + Tube & 0.9550 & 3.392  & 3.55\\
        2 & TCN + Tube & 0.9581 & 3.453 & 3.60\\
        3 & TCN + QD & 0.9589 & 3.511 & 3.66\\
        4 & LSTM + Tube & 0.9589 & 3.697 & 3.86\\
        5 & GRU + QD & 0.9666 & 3.966 & 4.10 \\
        6 & TCN + Quantile & 0.9674 & 4.094 & 4.23 \\
        7 & LSTM + QD & 0.9689 &4.470 & 4.61 \\
        8 & MDN  & 0.955 & 4.626 & 4.84 \\
        9 & LSTM + Quantile & 0.979 & 4.937 & 5.04 \\
        10 & GRU + Quantile & 0.9891 & 5.365 & 5.43 \\
        11 & Deep AR& 0.9969 &6.3553 & 6.38 \\
        12 & Time GPT & 0.9417 & 10.608 & 11.27 \\
        \hline
    \end{tabular}
\end{table}
We rank all used probabilistic forecasting models trained on the Jaisalmer dataset using our evaluation rule in Table \ref{tab:jaisalmer_ranking}. It places the Tube loss based deep models as the top performing probabilistic forecasting models on the Jaisalmer dataset. Figure \ref{fig:GRU_Tube} shows the forecasting of the top performing GRU+Tube model on Jaisalmer dataset.

\textbf{Los Angeles :-} On the Los Angeles dataset, for the LSTM architecture, the Tube loss clearly outperforms both the QD and Quantile approaches. For the GRU architecture, the Quantile approach achieves a marginally lower MPIW than both the QD and Tube loss functions. For the TCN architecture, the Tube loss and QD loss perform similarly, both outperforming the Quantile approach. However, the MDN model outperforms all the probabilistic forecasting models used on the Los Angeles dataset.
\begin{table}[h]
    \centering
    \caption{Los Angeles Dataset Rankings}
    \label{tab:losangeles_ranking}
    \begin{tabular}{|c|l|c|c|c|}
        \hline
        \textbf{Rank} & \textbf{Model} & \textbf{PICP} & \textbf{MPIW} & \textbf{MPIW/PICP}\\
        \hline
        1 & MDN & 0.956 & 3.487 & 3.65 \\
        2 & TCN + QD & 0.9507 & 3.548 & 3.73\\
        3 & TCN + Tube & 0.9502 & 3.555 & 3.74 \\
        4 & LSTM + Tube & 0.9553 & 3.589 & 3.76 \\
        5 & GRU + Quantile & 0.951 & 3.618 & 3.81 \\
        6 & GRU + Tube & 0.9586 & 3.737 & 3.90 \\
        7 & GRU + QD & 0.9528 & 3.749 & 3.93 \\
        8 & LSTM + Quantile & 0.9547 & 4.001 & 4.19 \\
        9 & Deep AR & 0.9773 & 4.5008 & 4.61 \\
        10 & Time GPT & 0.9517 & 6.519 & 6.85 \\
        11 & LSTM + QD & 0.9467 & 3.471 & 3.67 \\
        12 & TCN + Quantile & 0.9416 & 3.779 & 4.01 \\
        \hline
    \end{tabular}
\end{table}
In Table \ref{tab:losangeles_ranking}, we rank all the probabilistic forecasting models trained on the Los Angeles dataset.

 \textbf{San Francisco :-}  On the San Francisco dataset, our Tube loss method once again outperforms all recently developed baseline methods. In Table \ref{tab:sanfrancisco}, we rank all the probabilistic forecasting models trained on the San Francisco dataset.
 \begin{table}[h]
    \centering
    \caption{San Francisco Dataset Rankings}
    \label{tab:sanfrancisco_ranking}
    \begin{tabular}{|c|l|c|c|c|}
        \hline
        \textbf{Rank} & \textbf{Model} & \textbf{PICP} & \textbf{MPIW} & \textbf{MPIW/PICP} \\
        \hline
        1 & TCN + Tube & 0.9543 & 4.560 & 4.78 \\
        2 & LSTM + Tube & 0.9561 & 4.627 & 4.84 \\
        3 & GRU + Tube & 0.9507 &4.857 & 5.11 \\
        4 & GRU + Quantile & 0.951 & 4.955 & 5.21 \\
        5 & LSTM + Quantile & 0.9594 & 5.407 &5.64 \\
        6 & TCN + QD & 0.9505 & 5.490 &5.78 \\
        7 & LSTM + QD & 0.9734 & 6.043 & 6.21 \\
        8 & Deep AR & 0.9855 & 6.2023 & 6.29 \\
        9 & GRU + QD & 0.9724 & 6.309 & 6.49 \\
        10& MDN & 0.949 & 4.620 & 4.87  \\
        11 & TCN + Quantile & 0.9428 & 4.908 & 5.21 \\
        12 & TimeGPT & 0.9357 & 11.235 & 12.00 \\
        \hline
    \end{tabular}
\end{table}

 The TCN+Tube model performs best, achieving a PICP of 0.9543 with an MPIW of 4.56 approximately 7\% narrower than TCN + Quantile (MPIW = 4.908). Likewise, LSTM + Tube obtains the highest PICP among Tube-based models at 0.9561, while keeping the MPIW at 4.627. This is around 30\% tighter than LSTM + QD, which, although yielding a higher PICP of 0.9734, has a much broader MPIW of 6.043.

\subsection{Conclusion}
Across all three datasets, the deep learning architectures based on Tube loss consistently demonstrate strong and reliable performance. Table \ref{tab:average_ranking1} presents the final mean ranks of all forecasting methods evaluated across the datasets. Notably, the top three positions are occupied by Tube loss-based models: TCN + Tube achieves the highest rank, followed by GRU + Tube and LSTM + Tube. Although TCN + QD ranks fourth in Table \ref{tab:average_ranking1}, the QD loss does not show consistent performance when combined with other deep architectures such as LSTM and GRU.

\begin{table}[h]
    \centering
    \caption{Average Model Rankings Across Datasets}
    \label{tab:average_ranking1}
    \begin{tabular}{|l|c|c|}
        \hline
        \textbf{Rank} & \textbf{Model} & \textbf{Average Rank} \\
        \hline
        1 & TCN + Tube & 2.00  \\
        2 & GRU + Tube & 3.33  \\
        3 & LSTM + Tube & 3.33  \\
        4 & TCN + QD & 3.67  \\
        5 & MDN & 6.33  \\
        6 & GRU + Quantile & 6.33  \\
        7 & GRU + QD & 7.00  \\
        8 & LSTM + Quantile & 7.33  \\
        9 & LSTM + QD & 8.33 \\
        10 & TCN + Quantile & 	9.67  \\
        11 & Deep AR & 	9.33  \\
        12 & TimeGPT & 11.33 \\
        \hline
    \end{tabular}
\end{table}

\begin{table}[h]
    \centering
    \caption{Average Improvement of Tube Loss with Other Losses}
    \label{tab:avg_improvement}
    \begin{tabular}{|l|c|c|}
        \hline
        \textbf{Dataset} & \textbf{Average MPIW} & \textbf{ $\%$ of Improvement} \\
        \hline
        \multicolumn{3}{|c|}{\textbf{San Francisco}} \\
        \hline
        Tube Loss & 4.681 & Baseline \\
        QD Loss & 5.947 & 21.29\% \\
        Quantile  & 5.181 & 9.65\% \\
        Deep AR & 6.2023 & 24.53\% \\
        \hline
        \multicolumn{3}{|c|}{\textbf{Jaisalmer}} \\
        \hline
        Tube Loss & 3.514 & Baseline \\
        QD Loss & 3.982 & 11.75\% \\
        Quantile  & 4.798 & 26.76\% \\
        Deep AR & 6.3553 & 44.71\% \\
        MDN & 4.626 & 24.04\% \\
        \hline
        \multicolumn{3}{|c|}{\textbf{Los Angeles}} \\
        \hline
        Tube Loss & 3.627 & Baseline \\
        QD Loss & 3.648 & 2.10\% \\
        Quantile  & 3.809 & 4.78\% \\
        Deep AR & 4.5008 & 19.41\% \\
        MDN & 3.487 & MDN is better by 4.014\% \\
        Time GPT & 6.519 & 44.36\% \\
        \hline
    \end{tabular}
\end{table}

\begin{figure}
\centering
\includegraphics[width=0.90\linewidth,height = 0.40\linewidth]{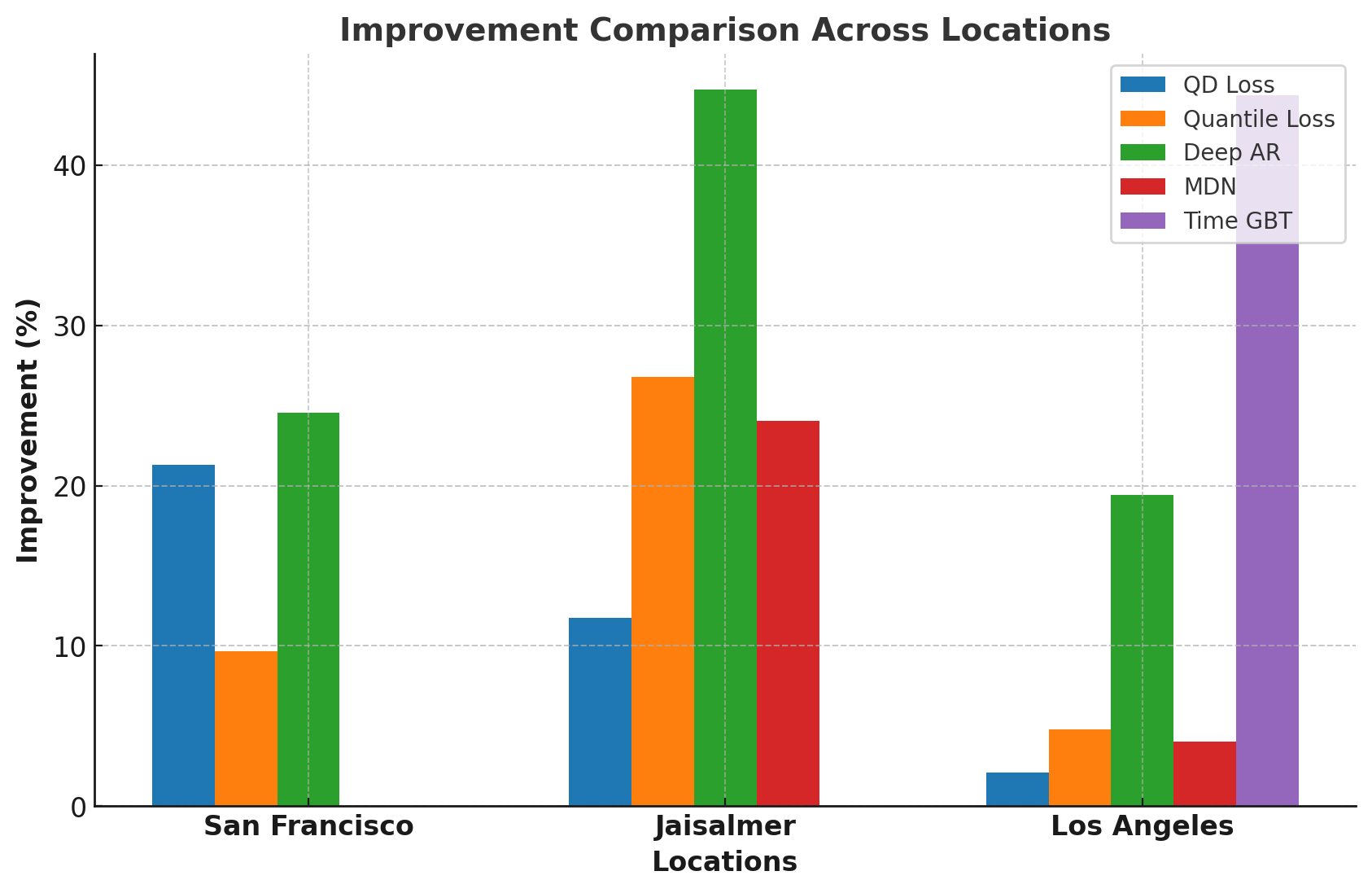}
\caption{Average Improvement obtained by the Tube loss model over existing forecasting models}
\label{fig:avg_rank}
\end{figure}

Tube loss-based models consistently achieve the target coverage of 0.95 across all scenarios, regardless of the underlying deep architecture. In contrast, the QD loss model fails to meet the 0.95 coverage target in one instance (LSTM+QD on the Los Angeles dataset), while the quantile-based model fails twice (TCN+Quantile on the Los Angeles and San Francisco datasets). Additionally, the quantile-based method sometimes results in coverage significantly higher than the target 0.95, which leads to increased MPIW. This is due to the absence of a mechanism within the model to explicitly minimize the width of the PI.

The MPIW quantifies the width of the obtained PI which should be narrow as possible to reduce the uncertainties in decisions. In Table \ref{tab:avg_improvement}, we list the average MPIW values obtained by deep learning models using the Tube, QD loss and Quantile approaches on all datasets along with the MDN and Deep AR methods. Clearly the Tube loss based deep architectures obtain the narrower PI as compared to the other deep learning models. We quantify this improvement by computing the  metric, "$\%$ of improvement"  which is computed by  ( MPIW of X - MPIW of Tube loss ) $\times$ 100 /  MPIW of X and listed them also in \ref{tab:avg_improvement}.  To visualize this percentage of the improvement across different datasets obtained by the Tube loss models, we have plotted the  Figure \ref{fig:avg_rank}. The Tube loss based deep forecasting method can obtain the significant improvement of the MPIW values as compared to the other existing model.

\bibliography{wind.bib}

\end{document}